\documentclass[runningheads]{llncs}
\usepackage[colorlinks]{hyperref}
\usepackage{tabularx}
\usepackage{amsmath}
\usepackage{amssymb}
\usepackage[utf8]{inputenc} 
\usepackage{graphicx}
\usepackage{arydshln}
\begin{document}
\title{Hybrid Deep Learning Gaussian Process for Diabetic Retinopathy Diagnosis and Uncertainty Quantification}
\titlerunning{Deep Learning Gaussian Process for Diabetic Retinopathy}
%
\author{ Santiago Toledo-Cortés\inst{1} \and
Melissa De La Pava\inst{1} \and
Oscar Perdómo \inst{2} \and
Fabio A. González \inst{1}}
\authorrunning{S. Toledo-Cortés et al.}
%
\institute{MindLab Research Group, Universidad Nacional de Colombia, Bogotá, Colombia \email{\{stoledoc, medel, fagonzalezo\}@unal.edu.co} \and Universidad del Rosario, Bogotá, Colombia \email{oscarj.perdomo@urosario.edu.co}}
\maketitle              
\begin{abstract}

  Diabetic Retinopathy (DR) is one of the microvascular complications of \emph{Diabetes Mellitus}, which remains as one of the leading causes of blindness worldwide. Computational models based on Convolutional Neural Networks represent the state of the art for the automatic detection of DR using eye fundus images. Most of the current work address this problem as a binary classification task. However, including the grade estimation and quantification of predictions uncertainty can potentially increase the robustness of the model. In this paper, a hybrid Deep Learning-Gaussian process method for DR diagnosis and uncertainty quantification is presented. This method combines the representational power of deep learning, with the ability to generalize from small datasets of Gaussian process models. The results show that uncertainty quantification in the predictions improves the interpretability of the method as a diagnostic support tool. The source code to replicate the experiments is publicly available at \url{https://github.com/stoledoc/DLGP-DR-Diagnosis}
 
\end{abstract}
\section{Introduction}

Diabetic Retinopathy (DR) is a consequence of \emph{Diabetes Mellitus} that manifests itself in the alteration of vascular tissue. When an alteration in the correct blood supply occurs, lesions such as microaneurysms, hemorrhages and exudates appear \cite{Wilkinson2003}.  These lesions can be identified in eye fundus images, one of the fastest and least invasive methods for DR diagnosing. 
Although early detection and monitoring are crucial to prevent progression and loss of vision \cite{Wells2016}, in developing countries approximately 40\% of patients are not diagnosed due to lack of access to the medical equipment and specialist, which puts patients of productive age at risk of visual impairment \cite{Wilkinson2003,Yau2012}. Therefore, to facilitate access to rapid diagnosis and speed up the work of professionals, many efforts have been made in the development of machine learning models focused on the analysis of eye fundus images for automatic DR detection. 

For medical image analysis, deep Convolutional Neural Networks represent the state of the art. These methods work by means of filters that go through the image and exploit the natural structure of the data, being able to detect increasingly complex patterns. However, the success of these deep learning models depends on the availability of very large volumes of data, and this is not always the case for medical image datasets. For instance, one of the largest public-available image dataset for DR detection is EyePACS \cite{kaggle}, which has 35126 samples for training. For this reason, training a deep learning model for this problem from scratch is not always feasible \cite{Mahmut}. Instead, fine-tuning of pretrained models is preferred, as it allows the models to refine a general knowledge for an specific tasks. However, the number of specific sample images is not always enough to make a tuning that produces good final performances \cite{Mahmut}.

Classical machine learning methods such as Gaussian Processes (GP), on the other hand, were originally designed to work well with small data sets \cite{Alpaydin}. They have different advantages over deep neural network models, as lower number of parameters to train, convex optimization, modularity in model design, the possibility to involve domain knowledge, and in the case of Bayesian approaches, they allow the calculation of prediction uncertainty \cite{Wilson}. The latter would be useful in medical applications, as it gives to the final user an indication of the quality of the prediction \cite{Leibig}.

This work presents and evaluates a hybrid deep learning-Gaussian process model for the diagnosis of DR, and prediction uncertainty quantification. Taking advantage of the representational power of deep learning, features were extracted using an Inception-V3 model, fine-tuned with EyePACS dataset. With these features we proceed to train a GP regression for DR grading.

Our framework shows that:

\begin{enumerate}
    \item The performance of the proposed hybrid model trained as a regressor for the DR grade, allows it to improve binary classification results when compared with the single deep learning approach.
    \item Gaussian processes can improve the performance of deep learning methods by leveraging their ability to learn good image representations, when applied for small datasets analysis.
    \item The integration of GP endows the method with the ability to quantify the uncertainty in the predictions. This improves the usability of the method as a diagnostic support tool. Furthermore the experimental results show that the predictions uncertainty is higher for false negatives and false positives than for true positives and true negatives respectively. This is a high valued skill in computational medical applications.
\end{enumerate}
 
The paper is organized as follows: Section 2 presents a brief review of the previous work related to the diagnosis and calculation of uncertainty of the of DR automatic classification. Section 3 introduce the theoretical framework for the experiments, which will be described in Section 4. Finally, in Section 5 the discussion of the results and conclusions are presented.

\section{Related Work}
Many approaches have been proposed for the DR binary detection, most of them based in deep neural networks \cite{Perdomo2019}. Some of them combine deep models with metric learning techniques, as in \cite{Zeng2019}, where an Inception-V3 is trained and embedded into siamese-like blocks. The final DR binary or grade prediction is given by a fully-connected layer. In \cite{Gargeya2017}, a customized deep convolutional neural network to extract features is presented. The features and multiple metadata related to the original fundus image are used to trained a gradient boosting classifier to perform the DR prediction. In \cite{Li2019} an Inception-V3 model is once again fine-tuned using a private set of eye fundus images, but not with binary labels, but with five DR grade labels. The results are reported using a subset of the Messidor-2 dataset \cite{Abramoff2013,Decenciere2014}. This makes performance comparison impossible with many other results, including those presented in this paper. Better results were reported by Gulshan et al. in \cite{Gulshan2016}, where an ensemble of ten Inception-V3 models, pretrained on ImageNet, are fine-tuned on a non-public eye fundus image dataset. The final classification is calculated as the linear average over the predictions of the ensemble. Results on Messidor-2 were reported, with a remarkable 99\% AUC score. In \cite{Voets}, Voets et al. attempted to reproduce the results presented in \cite{Gulshan2016}, but it was not possible since the original study used non-public datasets for training. However, Voets et al. published the source code and models, and details on training procedure and hyperparameters are published in \cite{Voets} and \cite{Krause2018}.




Regarding the estimation of predictive uncertainty, the first work in this matter in DR detection models was proposed in \cite{Leibig}, where bayesian inference is used for uncertainty quantification in binary classification of DR. Another approach is presented in \cite{Lim2018}, where stochastic batch normalization is used to calculate the uncertainty of the prediction of a model for DR level intervals estimation. In the work presented in \cite{Raghu2019}, a dataset with multiple labels given by different doctors for each patient is used, which allows the calculation of uncertainty to predict professional disagreement in a patient diagnosis.

In relation to convolutional neural networks uncertainty estimation using GP, some work has been done specially outside the DR automatic detection context, as in \cite{Xin2019}, where a framework is developed to estimate uncertainty in any pretrained standard neural network, by modelling the prediction residuals with a GP. This framework was applied to the IMDB dataset, for age estimation based in face images. Also in \cite{Bradshaw2017}, a GP on the top of a neural networks is end-to-end trained, which makes the model much more robust to adversarial examples. This model was used for classification in the MNIST and CIFAR-10 datasets.

To our knowledge, this is the first work that implements a GP to quantify the uncertainty of a model predictions of DR diagnosis. 

\section{Deep Learning Gaussian Process For Diabetic Retinopathy Diagnosis (DLGP-DR)}

\begin{figure}
    \centering
    \includegraphics[width=\textwidth]{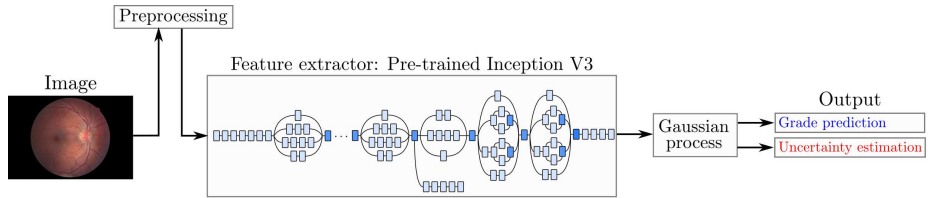}
    \caption{Proposed DLGP-DR model. Fine-tuned Inception-V3 is used as feature extractor. The extracted features are then used to train a Gaussian process.}
    \label{fig_method}
\end{figure}

The overall strategy of the proposed Deep Learning Gaussian Process For Diabetic Retinopathy grade estimation (DLGP-DR) method comprises three phases, and is shown in Fig.~\ref{fig_method}. The first phase is a pre-processing stage, described in \cite{Voets}, which is applied to all eye fundus image datasets used in this work. This pre-processing eliminates the very dark images where the circular region of interest is not identified, eliminates the excess of black margin, and resizes the images to 299$\times$299 pixels. The second phase is a feature extraction. An Inception-V3 model, trained with ImageNet and fine-tuned with EyePACS dataset is used as feature extractor. Each sample is then represented by a 2048-dimensional vector. The third and final task is the DR diagnosis, which is performed by a GP regressor.

\subsection{Feature extraction - Inception-V3}

Many previous works have used deep learning models for the diagnostic of DR. Recently, Voets et al. \cite{Voets} attempted to replicate the results published in \cite{Gulshan2016}, by fine-tuning an assembly of ten pretrained Inception-V3 networks. While Voets et al. were not able to achieve the same results reported in \cite{Gulshan2016}, most of the implementation details, as well as the specific partitioning for the training and test sets are publicly accessible, and were used in this study in the fine tuning of an Inception-V3 model. Once trained, the feature extraction is achieved by defining the global average pooling layer of the network as the output of the model, and use it to predict all the images in the datasets. Thus, each image will be represented by 2048 features which are used to train and evaluate the GP model. 


\subsection{Gaussian Processes}

Gaussian processes are a Bayesian machine learning regression approach that are able to produce, in addition to the predictions, a value of uncertainty about them \cite{Alpaydin}. The method requires as input a covariance given by a kernel matrix. The kernel matrix would be the gram matrix computed over the training set with a Radial Basis Function (RBF). This RBF kernel depends on two parameters which will be learned during the training process. We performed a Gaussian process regression, where the labels are the five grades of retinopathy present in the EyePACS dataset. From the prior, the GP calculates the probability distribution of all the functions that fit the data, adjusting the prior from the evidence, and optimizing the kernel parameters. Predictions are obtained by marginalizing the final learned Gaussian distribution, which in turn yields another normal distribution, whose mean is the value of the prediction, and its standard deviation gives a measure of the uncertainty of the prediction. Thus, an optimized metric (attached to a RBF similarity measure) is learned from the data, used to estimate the DR grade.

This GP can be adapted to do binary classification. One simple way to do this is defining a linear threshold in the prediction regression results.  The standard way however, consist in training a GP with binary labels and filtering the output of the regression by a sigmoid function. This results in a Gaussian Process Classifier (GPC). In any case, the predictions of a GPC are not longer subject to a normal distribution, and the uncertainty can not be measured. Therefore, the GPC will not take part in this study.

\section{Experimental evaluation}


\subsection{Datasets}

Experiments were performed with two eye fundus image datasets: EyePACS and Messidor-2. EyePACS comes with labels for five grades of DR: grade 0 means no DR, 1, 2, and 3 means non-proliferative mild, moderate and severe DR, while grade 4 means proliferative DR. For the binary classification task, according to the International Clinical Diabetic Retinopathy Scale \cite{severity_2002}, grades 0 and 1 corresponds to non-referable DR, while grades 2, 3, and 4 correspond to referable DR.  In order to achieve comparable results with \cite{Voets}, we took the same EyePACS partition used for training and testing (see Table \ref{tab1}). This partition was constructed only to ensure that the proportion of healthy and sick examples in training and testing was the same as that reported in \cite{Gulshan2016}. EyePACS train set is used for training and validation of the Inception-V3 model. Then, the feature extraction described in Section 3 is applied. The extracted features are used for training the DLGP-DR model. The evaluation is performed on the EyePACS test set and on the Messidor-2, which is a standard dataset used to compare performance results in DR diagnosis task. Datasets details are described in Table~\ref{tab1} and in Table~\ref{tab2}.

\begin{table}
\begin{center}
\caption{Details of Messidor-2 dataset used for testing. Class 0 correspond to non-referable cases.}\label{tab1}
\begin{tabular}{|c|c|c|c|}
\hline
 ~Class~ & ~Test Samples~\\
\hline
  0  & 1368\\
  1  & 380\\
  
\hline

\end{tabular}
\end{center}
\end{table}

\begin{table}
\begin{center}
\caption{Details of the subset and final partition of the EyePACS dataset used for training and testing. This is the same partition used in \cite{Voets}. Grades 0 and 1 correspond to non-referable patients, while grades 2, 3, and 4 correspond to referable cases.}\label{tab2}
\begin{tabular}{|c|c|c|c|}
\hline
 ~Grade~ & ~ Train Samples~ & ~Test Samples~\\
\hline
  0 & 37209 & 7407\\
  1 & 3479 & 689\\
  2 & 12873 & 0\\
  3 & 2046  & 0\\
  4 & 1220  & 694\\
\hline

\end{tabular}
\end{center}
\end{table}

\subsection{Experimental Setup}

Fine-tuning was made to an Inception-V3 network, pretrained on ImageNet and available in Keras \cite{Szegedy2016}. The model was trained for binary DR classification task. The data augmentation configuration for horizontal reflection, brightness, saturation, hue, and contrast changes, is described in \cite{Krause2018}, and it is the same used in \cite{Voets} and in \cite{Gulshan2016}. The top layer of the Inception-V3 model is removed and replaced by two dense layers of 2048 and 1 neurons. BinaryCrossentropy was used as loss function and RMSprop as optimizer, with a learning rate of $10^{-6}$ and a decay of $4\times10^{-5}$. The performance of the model is validated by measuring the AUC in a validation set consisting of 20\% of the training set. 

Once the model is trained, the average pooling layer from the Inception-V3 model is then used as output for feature extraction. The extracted features from the Inception-V3 are normalized and used to train a GP regressor over the five DR grade labels, it means, to perform the DR grading task. Therefore, the output of the DLGP-DR is a continuous number indicating the DR grade. 

Two baselines were defined to compare the DLGP-DR performance. Results reported by Voets es al. \cite{Voets} constitute the first baseline of this study. The second baseline is an extension of the Inception-V3 model with two dense layers trained on the same feature test as the Gaussian process, which is called as \textit{NN-model} hereafter.

\subsection{EyePACS results}

\begin{figure}
\centering
\begin{minipage}[b]{0.48\textwidth}
\includegraphics[width=\textwidth]{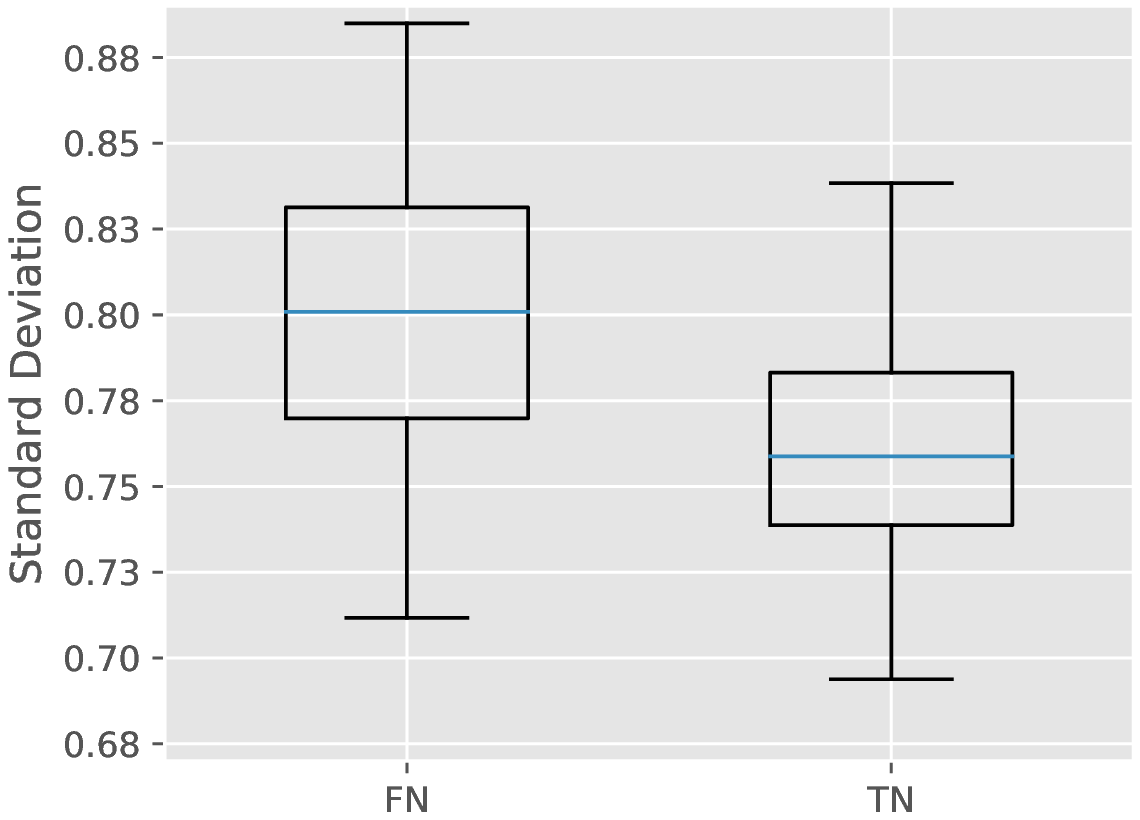}
\caption{Standard deviation for samples predicted as negative (non-referable) instances by DLGP-DR. FN: false negatives, TN: true negatives.}
\label{fig:y_y_ped}
\end{minipage}
\hfill
\begin{minipage}[b]{0.48\textwidth}
\includegraphics[width=\textwidth]{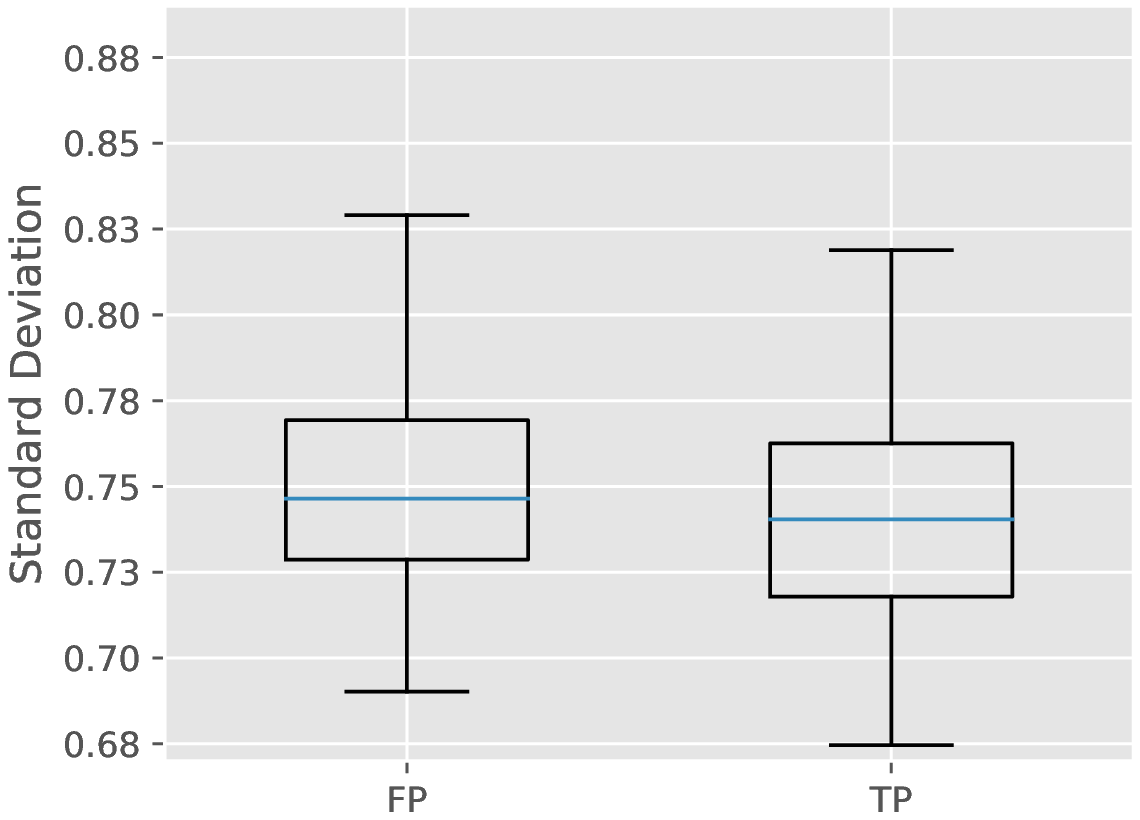}
\caption{Standard deviation for samples predicted as positive (referable) instances by DLGP-DR. FP: false positives, TP: true positives.}
\label{fig:std_fp_fn}
\end{minipage}
\end{figure}

DLGP-DR is evaluated in the EyePACS test partition. The results are binarized using a threshold of 1.5 (which is coherent with referable DR detection), and compared with baselines in Table~\ref{tab3}. In addition, although uncertainty estimation is not used to define or modify the prediction, DLGP-DR uncertainty is analysed for false positives (FP), false negatives (FN), true positives (TP) and true negatives (TN). As mentioned before, referable diabetic retinopathy is defined as the presence of moderate, severe and proliferative DR. So, the false negatives are calculated as the patients that belong to grade 4 but are classified as grades 0 and 1. The false positives are calculated as the patients belonging to grades 0 and 1 but classified in grade 4. The results are shown in Fig.~\ref{fig:y_y_ped} and Fig.~\ref{fig:std_fp_fn}.



\subsection{Messidor-2 results}

For Messidor-2 dataset, the predictions given by DLGP-DR are binarized using athe same threshold of 1.5 used for EyePACS. Based on the results of the uncertainty measured in the EyePACS test dataset, those samples predicted negative for which the standard deviation was higher than 0.84, were changed to positive. The results are reported and compared with the baselines in the Table~\ref{tab4}.

\begin{table}[]
\begin{center}
\caption{Comparison performance of DLGP-DR for binary classification in EyePACS test partition used in \cite{Voets}. As it is not the standard EyePACS test set, comparison is not feasible with other similar studies.}\label{tab3}
\begin{tabular}{|c|c|c|c|}
\hline
Description & Sensitivity & Specificity & AUC \\
\hline
Voets 2019 \cite{Voets}  & 0.906 & 0.847  & 0.951 \\

\textit{NN-model} & 0.9207 & 0.85 & 0.9551\\
\hdashline
\textbf{DLGP-DR}  & \textbf{0.9323} & \textbf{0.9173} & \textbf{0.9769} \\
\hline
\end{tabular}
\end{center}
\end{table}

\begin{table}[]
\begin{center}
\caption{Comparison performance of DLGP-DR for binary classification in Messidor-2. Referenced results from \cite{Voets} were directly extracted from the respective documents.}\label{tab4}
\begin{tabular}{|c|c|c|c|}
\hline
Description & Sensitivity & Specificity & AUC \\
\hline
Voets 2019 \cite{Voets}  & \textbf{0.818} & 0.712  & 0.853 \\
\textit{NN-model} & 0.7368 & 0.8581 & 0.8753 \\
\hdashline
\textbf{DLGP-DR}  & 0.7237 & \textbf{0.8625} & \textbf{0.8787} \\
\hline
\end{tabular}
\end{center}
\end{table}

\subsection{Discussion}

Results reported in Table~\ref{tab3} shows that DLGP-DR outperforms specificity and AUC score of the \textit{NN-model} and outperforms all the metrics reported by Voets et al. \cite{Voets}. As observed in Table~\ref{tab4}, DLGP-DR outperforms both baselines for specificity and AUC scores. Although Gulshan et al. have reported $0.99$ for AUC score in Messidor-2 \cite{Gulshan2016}, as Voets et al. comments in \cite{Voets}, the gap in the results may be due to the fact that the training in that study was made with other publicly available images and with a different gradation made by ophthalmologists. Overall, this shows that the global performance of the DLGP-DR exceeds that of a neural network-based classifier. In addition, in Fig.~\ref{fig:y_y_ped} and Fig.~\ref{fig:std_fp_fn} the box-plot shows that the standard deviation is higher for false positives and false negatives. This means, that the DLGP-DR model has bigger uncertainties for wrong classified patients than for well classified. which provides the user a tool to identify wrong predictions. This behavior is especially visible for false negatives, which is the most dangerous mistake in medical applications, because a ill patient can leave out without a needed treatment. 

\section{Conclusions}

In this study we took a deep learning model fine-tuned on the EyePACS dataset as feature extractor. The final task of DR classification and grading was carried out by means of a Gaussian process. For DR binary classification, the proposed DLGP-DR model reached better results than the original deep learning model. We also showed that a fine DR grade classification improve the binary classification performance of the original model.

Also, the DLGP-DR enables an uncertainty analysis. This analysis showed that the model could allow the identification of both, false negatives and false positives. The former are important due to the high cost of classifying a patient as healthy when it is not, and the later because they increase the costs of health care. The comparison between the Gaussian process and a neural network classifier for DR grades, showed once again that Gaussian processes are better tools for the analysis of medical images, for which datasets are usually far small to be analyzed entirely with deep learning techniques.

Overall, we demonstrate that the integration of deep learning and classical machine learning techniques is highly feasible in applications with small datasets, taking advantage of the representational power of deep learning and the theoretical robustness of classical methods.

\bigskip

\textbf{Acknowledgments}

\bigskip

This work was partially supported by a Google Research Award and by the Colciencias project number 1101-807-63563. 

%
%
%
\bibliographystyle{splncs04}
%

\bibliography{bib}

\begin{thebibliography}{10}
\providecommand{\url}[1]{\texttt{#1}}
\providecommand{\urlprefix}{URL }
\providecommand{\doi}[1]{https://doi.org/#1}

\bibitem{Abramoff2013}
Abr{\`{a}}moff, M.D., Folk, J.C., Han, D.P., Walker, J.D., Williams, D.F.,
  Russell, S.R., Massin, P., Cochener, B., Gain, P., Tang, L., Lamard, M.,
  Moga, D.C., Quellec, G., Niemeijer, M.: {Automated analysis of retinal images
  for detection of referable diabetic retinopathy}. JAMA Ophthalmology
  \textbf{131}(3),  351--357 (2013). \doi{10.1001/jamaophthalmol.2013.1743}

\bibitem{severity_2002}
{American Academy of Ophthalmology}: {International clinical diabetic
  retinopathy disease severity scale detailed table}. International Council of
  Ophthalmology  (2002)

\bibitem{Bradshaw2017}
Bradshaw, J., Matthews, A.G.d.G., Ghahramani, Z.: {Adversarial Examples,
  Uncertainty, and Transfer Testing Robustness in Gaussian Process Hybrid Deep
  Networks}. \href{https://arxiv.org/pdf/1707.02476.pdf}{\tt
  arXiv:1707.02476v1} eprint pp. 1--33 (2017)

\bibitem{Decenciere2014}
Decenci{\`{e}}re, E., Zhang, X., Cazuguel, G., La{\"{y}}, B., Cochener, B.,
  Trone, C., Gain, P., Ord{\'{o}}{\~{n}}ez-Varela, J.R., Massin, P., Erginay,
  A., Charton, B., Klein, J.C.: {Feedback on a publicly distributed image
  database: The Messidor database}. Image Analysis and Stereology
  \textbf{33}(3),  231--234 (2014). \doi{10.5566/ias.1155}

\bibitem{kaggle}
{Diabetic Retinopathy Detection of Kaggle}: Eyepacs challenge.
  \url{www.kaggle.com/c/diabetic-retinopathy-detection/data}, accessed:
  2019-10-15

\bibitem{Alpaydin}
Ethem, A.: Introduction to Machine Learning. The MIT Press, 3 edn. (2014)

\bibitem{Gargeya2017}
Gargeya, R., Leng, T.: {Automated Identification of Diabetic Retinopathy Using
  Deep Learning}. Ophthalmology  \textbf{124}(7),  962--969 (2017)

\bibitem{Gulshan2016}
Gulshan, V., Peng, L., Coram, M., Stumpe, M.C., Wu, D., Narayanaswamy, A.,
  Venugopalan, S., Widner, K., Madams, T., Cuadros, J., Kim, R., Raman, R.,
  Nelson, P.C., Mega, J.L., Webster, D.R.: {Development and validation of a
  deep learning algorithm for detection of diabetic retinopathy in retinal
  fundus photographs}. JAMA - Journal of the American Medical Association
  \textbf{316}(22),  2402--2410 (2016)

\bibitem{Mahmut}
Kaya, M., Bilge, H.: Deep metric learning: A survey. Symmetry  \textbf{11},
  ~1066 (08 2019). \doi{10.3390/sym11091066}

\bibitem{Krause2018}
Krause, J., Gulshan, V., Rahimy, E., Karth, P., Widner, K., Corrado, G.S.,
  Peng, L., Webster, D.R.: {Grader Variability and the Importance of Reference
  Standards for Evaluating Machine Learning Models for Diabetic Retinopathy}.
  Ophthalmology  \textbf{125}(8),  1264--1272 (2018).
  \doi{10.1016/j.ophtha.2018.01.034}

\bibitem{Leibig}
Leibig, C., Allken, V., Ayhan, M.S., Berens, P., Wahl, S.: {Leveraging
  uncertainty information from deep neural networks for disease detection}.
  Scientific Reports  \textbf{7}(1),  1--14 (2017)

\bibitem{Li2019}
Li, F., Liu, Z., Chen, H., Jiang, M., Zhang, X., Wu, Z.: {Automatic detection
  of diabetic retinopathy in retinal fundus photographs based on deep learning
  algorithm}. Translational Vision Science and Technology  \textbf{8}(6)
  (2019). \doi{10.1167/tvst.8.6.4}

\bibitem{Lim2018}
Lim, Z.W., Lee, M.L., Hsu, W., Wong, T.Y.: {Building Trust in Deep Learning
  System towards Automated Disease Detection}. The Thirty-First AAAI Conference
  on Innovative Applications of Artificial Intelligence pp. 9516--9521 (2018)

\bibitem{Perdomo2019}
Perdomo, O., Gonzalez, F.: {A Systematic Review of Deep Learning Methods
  Applied to Ocular Images}. Ciencia e Ingenieria Neogranadina  \textbf{30}(1)
  (2019)

\bibitem{Raghu2019}
Raghu, M., Blumer, K., Sayres, R., Obermeyer, Z., Kleinberg, R., Mullainathan,
  S., Kleinberg, J.: Direct uncertainty prediction for medical second opinions.
  In: Proceedings of the 36th International Conference on Machine Learning.
  PMLR 97. Long Beach, California (2019)

\bibitem{Szegedy2016}
Szegedy, C., Vanhoucke, V., Ioffe, S., Shlens, J., Wojna, Z.: {Rethinking the
  Inception Architecture for Computer Vision}. Proceedings of the IEEE Computer
  Society Conference on Computer Vision and Pattern Recognition
  \textbf{2016-December},  2818--2826 (2016). \doi{10.1109/CVPR.2016.308}

\bibitem{Voets}
Voets, M., M{\o}llersen, K., Bongo, L.A.: {Reproduction study using public data
  of: Development and validation of a deep learning algorithm for detection of
  diabetic retinopathy in retinal fundus photographs}. PLoS ONE
  \textbf{14}(6),  1--11 (2019)

\bibitem{Wells2016}
Wells, J.A., Glassman, A.R., Ayala, A.R., Jampol, L.M., Bressler, N.M.,
  Bressler, S.B., Brucker, A.J., Ferris, F.L., Hampton, G.R., Jhaveri, C.,
  Melia, M., Beck, R.W.: {Aflibercept, Bevacizumab, or Ranibizumab for Diabetic
  Macular Edema Two-Year Results from a Comparative Effectiveness Randomized
  Clinical Trial}. Ophthalmology  \textbf{123}(6),  1351--1359 (2016)

\bibitem{Wilkinson2003}
Wilkinson, C.P., Ferris, F.L., Klein, R.E., Lee, P.P., Agardh, C.D., Davis, M.,
  Dills, D., Kampik, A., Pararajasegaram, R., Verdaguer, J.T., Lum, F.:
  {Proposed international clinical diabetic retinopathy and diabetic macular
  edema disease severity scales}. Ophthalmology  \textbf{110}(9),  1677--1682
  (2003)

\bibitem{Wilson}
Wilson, A., Nickisch, H.: Kernel interpolation for scalable structured gaussian
  processes (kiss-gp). In: Proceedings of the 32nd International Conference on
  Machine Learning. JMLR: W\&CP. Lille, France (2015)

\bibitem{Xin2019}
Xin, Q., Elliot, M., Miikkulainen, R.: {Quantifying Point-Prediction
  Uncertainty in Neural Networks via Residual Estimation with an I/O Kernel.}
  In: ICLR 2020. pp. 1--17. Addis Ababa, Ethiopia (2019)

\bibitem{Yau2012}
Yau, J.W., Rogers, S.L., Kawasaki, R., Lamoureux, E.L., Kowalski, J.W., Bek,
  T., Chen, S.J., Dekker, J.M., Fletcher, A., Grauslund, J., Haffner, S.,
  Hamman, R.F., Ikram, M.K., Kayama, T., Klein, B.E., Klein, R., Krishnaiah,
  S., Mayurasakorn, K., O'Hare, J.P., Orchard, T.J., Porta, M., Rema, M., Roy,
  M.S., Sharma, T., Shaw, J., Taylor, H., Tielsch, J.M., Varma, R., Wang, J.J.,
  Wang, N., West, S., Zu, L., Yasuda, M., Zhang, X., Mitchell, P., Wong, T.Y.:
  {Global prevalence and major risk factors of diabetic retinopathy}. Diabetes
  Care  \textbf{35}(3),  556--564 (2012)

\bibitem{Zeng2019}
Zeng, X., Chen, H., Luo, Y., Ye, W.: {Automated diabetic retinopathy detection
  based on binocular siamese-like convolutional neural network}. IEEE Access
  \textbf{7}(c),  30744--30753 (2019)

\end{thebibliography}

\end{document}